\documentclass[10pt,twocolumn,letterpaper]{article}

\usepackage{cvpr}

\definecolor{cvprblue}{rgb}{0.21,0.49,0.74}
\usepackage[pagebackref,breaklinks,colorlinks,allcolors=cvprblue]{hyperref}

\def\paperID{*****}
\def\confName{EgoVis @ CVPR}
\def\confYear{2026}

\title{Cross-Modal Action Recognition in Egocentric Video Using Mamba:\\
Integrating RGB and Hand Skeleton Streams via CLS Token Fusion Strategies}

\author{Juan Ignacio Bustos Gorostegui$^{1,2}$ \quad\quad Maria Elena Buemi$^{1,2}$ \\[0.5em]
$^{1}$Univ. of Buenos Aires. Faculty of Exact and Natural Sciences. Dept. of Computer Science (DC). \\
$^{2}$CONICET-Univ. of Buenos Aires. Institute of Computer Sciences (ICC). Argentina. \\
}

\begin{document}
\maketitle

\begin{abstract}
Egocentric action recognition is a challenging task due to erratic camera motion, frequent hand occlusion, and the difficulty of maintaining consistent visual representations over time. In this work, we propose a cross-modal architecture that combines RGB video and temporal hand skeleton data within a unified Mamba-based framework, exploiting the linear time complexity of State Space Models (SSMs). 
Our architecture consists of three components: a VideoMamba module for visual feature extraction, a skeleton encoder built on a stack of Mamba blocks, and a fusion module that integrates both modalities into a single representation. A central contribution of this work is the design and evaluation of four Class (CLS) token mixing strategies for multimodal fusion: Naive, Average, Weighted and Context-based. These strategies differ in how the pretrained unimodal CLS tokens, which role is to act as information sinks concentrating learned representations, are leveraged to initialize the mixed CLS token used for final classification. We evaluate all strategies on the H2O dataset. 
Experimental results show that the Average strategy achieves the best performance, yielding gains of over 10\% Top-1 accuracy in the Tiny configuration and 25\% in the Small configuration over the VideoMamba baseline.
\end{abstract}

\section{Introduction}

Egocentric videos face several challenges that fixed cameras typically avoid, such as erratic camera motion, frequent occlusion by the hands, and the possibility of temporarily losing visual contact with the main action. For this reason, a flexible architecture capable of maintaining a global video representation while being continuously updated with noisy or misleading frames is required. These challenges make Transformers ~\cite{vaswani2023attentionneed} particularly well suited for egocentric video understanding, as their self-attention mechanism enables the construction of strong spatio-temporal representations and allows the model to retain important visual information over long sequences, even in the presence of extended periods of misleading input.

In addition, the ability of Transformers to incorporate multimodal data into their representations allows skeletal motion information to be used to improve action prediction. Temporal hand skeleton data not only helps mitigate occlusion issues, such as when fingers are hidden behind objects or the hand itself, but also provides clean motion cues related to the performed action, thereby enhancing the overall understanding. This has become increasingly easier with the growing availability of egocentric datasets that provide synchronized hand skeleton annotations alongside the video stream~\cite{fu2025gigahandsmassiveannotateddataset, kwon2021h2o, zhan2024oakink2datasetbimanualhandsobject}, most notably the H2O and OakInkV2 dataset~\cite{kwon2021h2o, zhan2024oakink2datasetbimanualhandsobject}, which both captures a diverse set of everyday bimanual interactions (pouring, cutting, assembling, etc) from a first-person perspective with 
ground-truth hand keypoint annotations, making it particularly well suited for evaluating action recognition models in realistic daily-life scenarios.

Although these properties make Transformers a natural fit for egocentric video understanding, unfortunately the self-attention mechanism scales quadratically with sequence length, which limits the feasibility of Transformers for online video understanding applications. For this reason, Mamba~\cite{gu2023mamba}, a State Space Model (SSM)-based architecture, has recently gained attention due to its linear time complexity and performance comparable to Transformers in NLP tasks. Extensions of Mamba to visual domains, such as Vision Mamba~\cite{zhu2024visionmambaefficientvisual} and Video Mamba~\cite{li2024videomamba}, have demonstrated promising results in image and video processing.

Although skeleton-based action recognition models using Mamba have been explored~\cite{wen2025actionmamba,dawood2024simba}, to the best of our knowledge, no prior work has integrated skeleton-based information with RGB streams within this framework. We address this gap by proposing a cross-modal Mamba architecture and systematically evaluating four strategies for fusing pretrained unimodal Class(CLS) token representations into a unified classification token.
\section{Cross-Modal Fusion Design}
\label{sec:design}
\subsection{Two-Branch Architecture}
\begin{figure}[t]
  \centering
  \includegraphics[width=\linewidth]{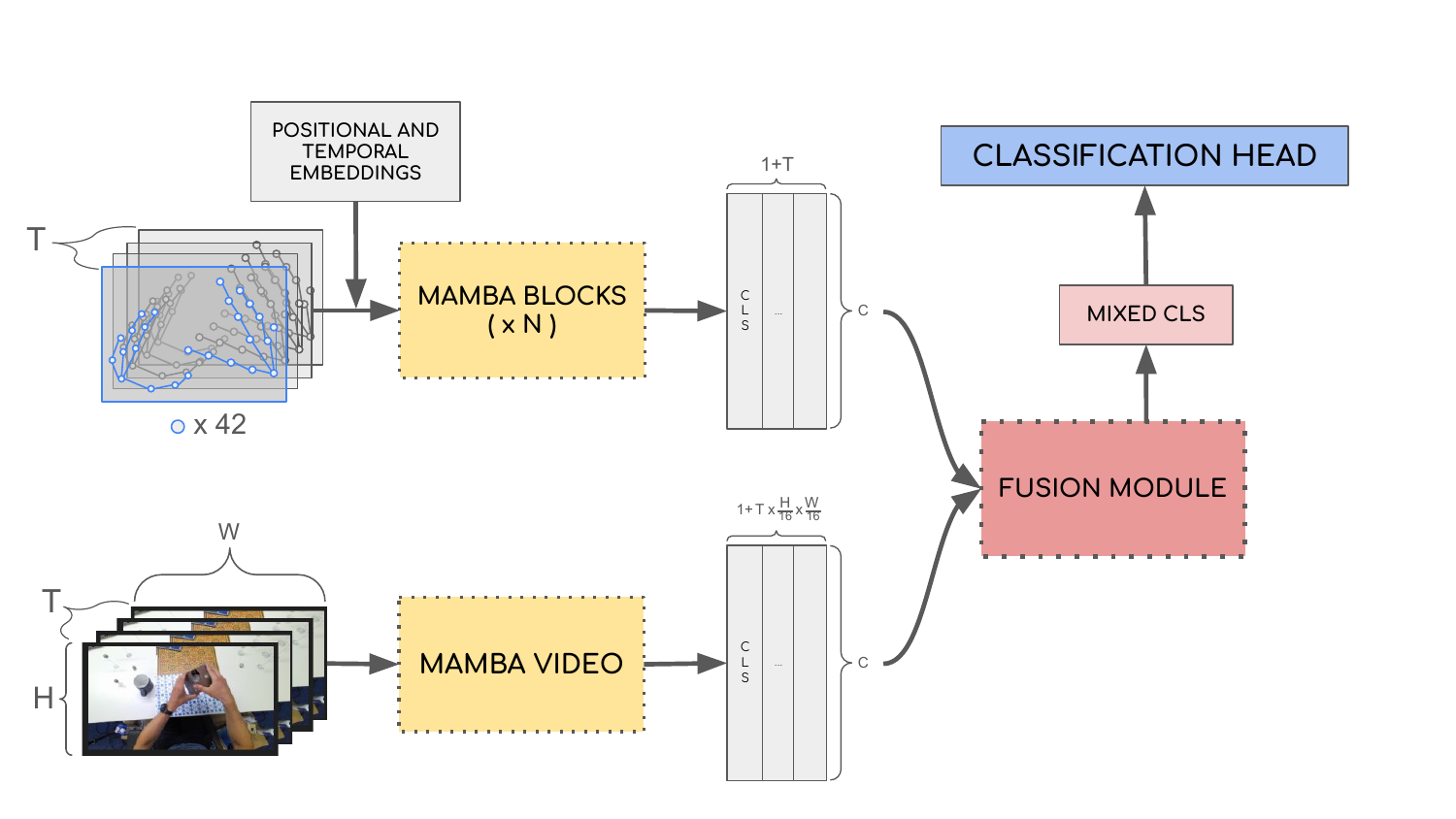}
  \caption{Overview of the proposed cross-modal architecture. The skeleton stream (top) and video stream (bottom) are encoded independently before being fused into a unified Mixed CLS token for classification. $T$ denotes the number of input frames, $H$ and $W$ the height and width of each frame respectively, and $C$ the embedding dimension. }
  \label{fig:architecture}
\end{figure}

The overall architecture is divided into three main components:
(1)~a VideoMamba module to extract embeddings from the video stream,
(2)~a pure Mamba module to encode skeleton data, and
(3)~a fusion module that combines both information streams into a unified representation, as can be seen in \cref{fig:architecture}.

Both the video and skeleton embeddings share the same dimensions. However, the number of tokens differs between modalities, as the video stream produces multiple tokens per frame due to spatial patchification.

The fusion module is inspired by the Mamba Suite Cross-Modal implementation~\cite{chen2024videomambasuite}. They add a learnable modality embedding to each modality-specific representation before concatenating them into a single sequence. Instead of keeping the original CLS tokens from each stream, they remove both and introduce a new one. The concatenated sequence is then processed through a Mamba block, and its final value is used for classification.

Our approach differs from the original implementation in how we handle the two modality-specific CLS tokens. In Mamba models, as in Transformers, it acts as an information sink, concentrating the relevant contextual information throughout processing before passing it to the classification head. By using pretrained video and skeleton encoders, their respective CLS tokens already contain dense and structured information. Discarding them entirely would waste this learned behavior.

\begin{figure}[t]
  \centering
  \includegraphics[width=\linewidth]{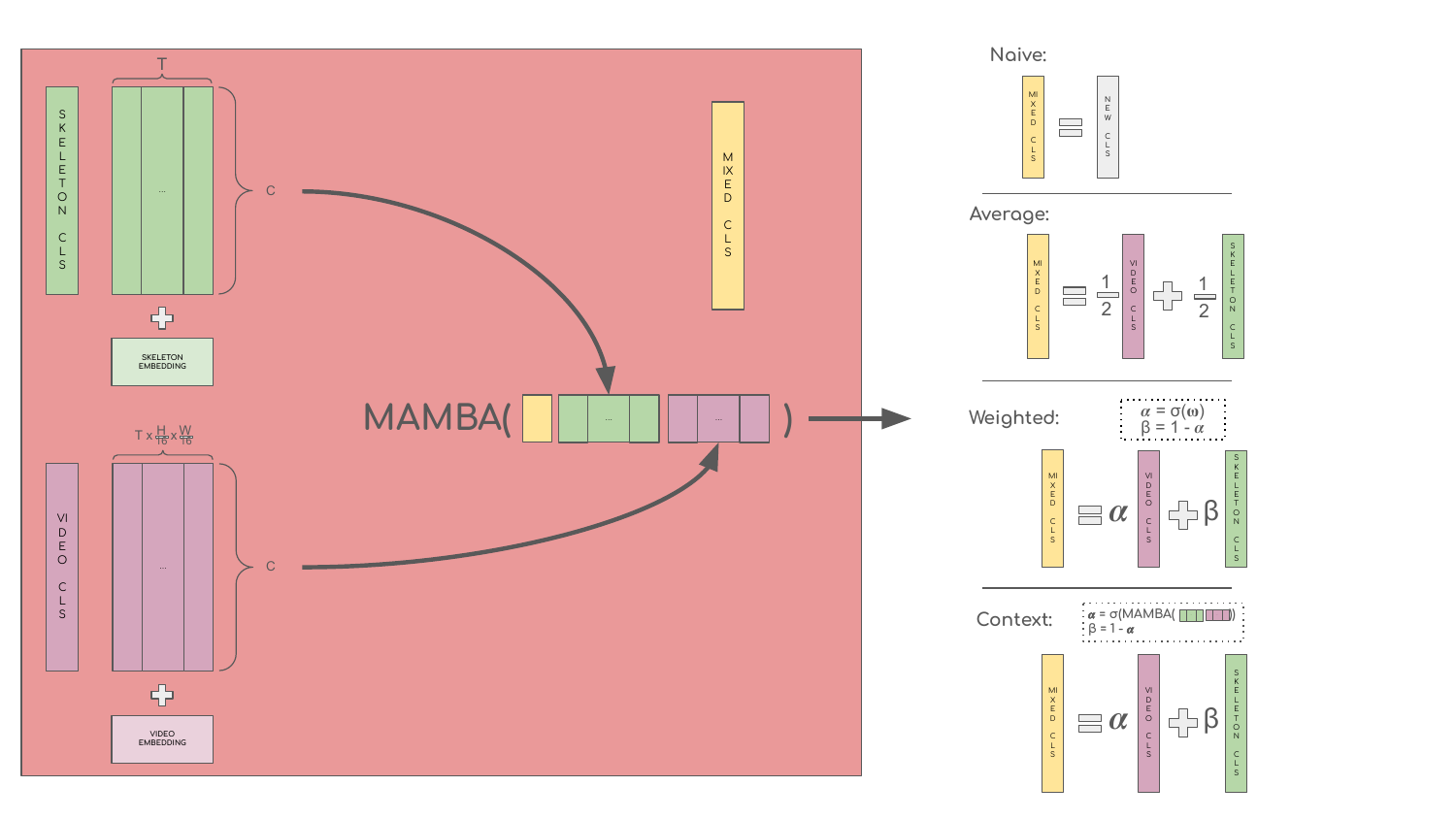}
  \caption{The four proposed CLS mixing strategies. From top to bottom: Naive (random initialization), Average (equal combination), Weighted (learnable scalar), and Context-based (input-dependent weight computed via a Mamba block). $C$ denotes the embedding dimension.}
  \label{fig:cls_strategies}
\end{figure}

For this reason, we propose four strategies to initialize the new CLS token, which we are referring to as the \emph{Mixed CLS token}, by leveraging the pretrained representations. A visual representation is shown in \cref{fig:cls_strategies}.

\subsection{CLS Mixing Strategies}

\noindent\textbf{Naive.} Both modality-specific CLS tokens are discarded, and a new trainable one is used.
\\

\noindent\textbf{Average.} The Mixed CLS token is initialized as the average by dimension of the video and skeleton CLS tokens.
\\

\noindent\textbf{Weighted.} A learnable scalar parameter $\omega$ is passed through a sigmoid function to compute a linear combination of the two CLS tokens:
\begin{equation}
    \text{CLS}_{mix} = \alpha\,\text{CLS}_{video} + (1 - \alpha)\,\text{CLS}_{skeleton},
    \label{eq:weighted}
\end{equation}
where $\alpha = \sigma(\omega)$.
\\

\noindent\textbf{Context-based.} Instead of learning a static scalar weight, we compute $\omega$ dynamically using a Mamba block that processes the concatenated token representations (excluding both CLS tokens). Specifically:
\begin{equation}
    \alpha = \sigma\!\left(\text{Mean}\!\left(\text{Mamba}(\mathbf{T}_{video} \oplus \mathbf{T}_{skel})   \right)\right),
    \label{eq:context}
\end{equation}
where $\mathbf{T}_{video}$ and $\mathbf{T}_{skel}$ denote the non-CLS token sequences from each modality, $\oplus$ denotes concatenation, and $\text{Mean}(\cdot)$ denotes taking the mean over the embedding values. The resulting $\alpha$ is then used in \cref{eq:weighted} to combine the two CLS tokens, allowing the model to adapt the fusion weight based on the input content.

\section{Experiments and Results}
\label{sec:experiments}

\subsection{Experimental Setup}

To evaluate the cross-modal capabilities of Mamba and the proposed fusion strategies, we conducted experiments on the H2O dataset~\cite{kwon2021h2o}, an egocentric video dataset that provides not only action labels but also ground-truth hand keypoints for each frame.

For both the VideoMamba module and the skeleton encoder, we adopted the same configurations defined in the VideoMamba model variants~\cite{li2024videomamba}, summarized in \cref{tab:configs}.

\begin{table}[ht]
\centering
\caption{Model configurations used for both the video and skeleton encoders. Depth denotes the number of stacked Mamba blocks in each branch.}
\label{tab:configs}
\begin{tabular}{lcc}
\hline
\textbf{Variant} & \textbf{Depth} & \textbf{Embedding Dim.} \\
\hline
Tiny   & 24 & 198 \\
Small  & 24 & 386 \\
\hline
\end{tabular}
\end{table}

The VideoMamba visual models were initialized from publicly available checkpoints pretrained for short-term video understanding on the Something-Something V2 dataset ~\cite{goyal2017somethingsomethingvideodatabase}, operating on 8-frame inputs. The skeleton encoder was trained directly on the H2O dataset.

We deliberately limited the fusion module to a single Mamba block, ensuring that any performance gains can be attributed to the model's ability to take advantage of the cross-modal information rather than to an increase in model depth.

Training was performed for 50 epochs using a linear warm-up schedule followed by a CosineAnnealingLR scheduler and the AdamW optimizer. Early stopping was applied if no validation improvement was observed for 10 consecutive epochs. All 12 models were trained on a single NVIDIA RTX 5090 GPU (32\,GB VRAM), with a total training time of approximately 2.5 hours.

\subsection{Analysis and Results}

As shown in \cref{tab:results_tiny}, the \textbf{Average} strategy achieves the best performance improvement, obtaining a gain of more than 10\% Top-1 accuracy over the VideoMamba baseline in the Tiny configuration and 25\% in the Small configuration. The \textbf{Weighted} strategy follows closely for the Tiny and Small variants, suggesting that the learnable parameter converges toward a balanced combination (\ie, $\alpha \approx 0.5$), effectively approximating the averaging strategy.

\begin{table}[htbp]
\centering
\caption{Top-1 accuracy (\%) on OakInkV2 for all fusion strategies compared against unimodal baselines. VideoMamba refers to the pretrained video encoder alone, and Skeleton Mamba to the skeleton-only Mamba stack. $\delta_V$ and $\delta_S$ denote absolute difference in accuracy points, while $\Delta_V$ and $\Delta_S$ denote relative change with respect to each baseline respectively.}
\label{tab:results_tiny}
\resizebox{\columnwidth}{!}{%
\begin{tabular}{lcccccc}
\hline
\multicolumn{6}{c}{\textbf{Tiny} (depth 24, dim 198)} \\
\hline
Method & Top-1 (\%) & $\delta_V$ & $\Delta_V$ & $\delta_S$ & $\Delta_S$ \\
\hline
Average                 & 61.90 & +5.71  & +10.17\% & +35.23 & +132.14\% \\
Weighted                & 58.10 & +1.90  & +3.39\%  & +31.43 & +117.86\% \\
\textbf{VideoMamba}     & \textbf{56.19} & --- & --- & +29.52 & +110.71\% \\
Naive                   & 50.48 & -5.71  & -10.17\% & +23.81 & +89.29\% \\
Context                 & 50.48 & -5.71  & -10.17\% & +23.81 & +89.29\% \\
\textbf{Skeleton Mamba} & \textbf{26.67} & -29.52 & -52.54\% & --- & --- \\
\hline
\multicolumn{6}{c}{\textbf{Small} (depth 24, dim 386)} \\
\hline
Method & Top-1 (\%) & $\delta_V$ & $\Delta_V$ & $\delta_S$ & $\Delta_S$ \\
\hline
Average                 & 60.95 & +12.38 & +25.49\% & +28.57 & +88.24\% \\
Context                 & 60.00 & +11.43 & +23.53\% & +27.62 & +85.29\% \\
Naive                   & 56.19 & +7.62  & +15.69\% & +23.81 & +73.53\% \\
Weighted                & 54.29 & +5.72  & +11.77\% & +21.90 & +67.65\% \\
\textbf{VideoMamba}     & \textbf{48.57} & --- & --- & +16.19 & +50.00\% \\
\textbf{Skeleton Mamba} & \textbf{32.38} & -16.19 & -33.33\% & --- & --- \\
\hline
\end{tabular}%
}
\end{table}

Inspection of the learned weight parameter $\alpha$ across model configurations reveals values of 0.62, 0.64 for the Tiny, Small variants respectively. These values consistently fall near 0.5, confirming that the weighted model converges toward a near-equal combination of both modalities. Notably, the slight but consistent bias above 0.5 suggests that the visual stream is marginally weighted higher than the skeleton stream, which is interpretable given that the VideoMamba encoder benefits from large-scale pretraining on Something-Something V2, while the skeleton encoder is trained exclusively on the comparatively small H2O dataset.

As expected, the \textbf{Naive} strategy consistently performs worse than the other fusion approaches. Since it discards both pretrained CLS tokens and replaces them with a fresh one, it does not effectively exploit the cross-modal information learned by the unimodal encoders.

Conversely, in the Small configuration, where both modalities contain meaningful information, the cross-modal model surpasses the video-only baseline even with the addition of only a single extra Mamba block. This demonstrates the complementary nature of the two modalities when both provide informative signals.

Contrary to expectations, the \textbf{Context-based} strategy does not outperform the simpler fusion methods, despite its higher flexibility. Whether this is a fundamental architectural limitation or a consequence of the limited scale of the datasets used remains an open question that warrants further investigation.

\section{Conclusions and Future Work}

In this work, we introduced a cross-modal Mamba-based architecture for egocentric action recognition that fuses RGB video streams with temporal hand skeleton data. We proposed and evaluated four CLS token fusion strategies (Naive, Average, Weighted, and Context-based) designed to leverage the dense, structured information encoded in pretrained unimodal CLS tokens when initializing a joint representation for classification.

Our experiments on the H2O dataset demonstrate that combining visual and skeletal modalities within a Mamba framework consistently outperforms the video-only baseline when both modalities carry meaningful information. The Average strategy proved to be the most effective fusion approach, achieving Top-1 accuracy gains of over 10\% and 25\% in the Tiny and Small model configurations respectively. The Weighted strategy performed comparably, likely converging toward a balanced combination equivalent to averaging. The Naive strategy, which discards pretrained CLS tokens entirely, consistently underperformed, confirming that preserving the learned unimodal representations during fusion is critical. Notably, the Context-based strategy, despite its greater flexibility, did not surpass simpler methods.

Several lines of work have originated from these results and are currently being pursued. First, evaluating the proposed strategies on larger egocentric datasets would clarify whether the saturation observed with the Context-based strategy is a data limitation or a fundamental architectural constraint. Second, an especially compelling direction is the use of skeleton data exclusively during training as a form of privileged information or cross-modal distillation, with the goal of improving visual-only inference without requiring skeleton inputs at test time. This would make the approach practical in deployment settings where hand keypoint estimation is unavailable or computationally prohibitive.

{
    \small
    \bibliographystyle{ieeenat_fullname}
    \bibliography{main}
}

\end{document}